\documentclass{ecai}
\usepackage{times}
\usepackage{graphicx}
\usepackage{latexsym}

\usepackage{algorithm}
\usepackage{algpseudocode}
\usepackage{amssymb}
\usepackage{amsmath}
\usepackage{setspace}
\usepackage[hyperfootnotes=false]{hyperref}

\usepackage{csquotes}
\usepackage{xcolor}
\usepackage[normalem]{ulem}

\usepackage{pifont}
\usepackage{caption}

\usepackage{EvolvingAI}
\newif\ifcomments

 \commentstrue

\ifcomments
\newcommand{\comments}[1]{#1}
\else
\newcommand{\comments}[1]{}
\fi

\newcommand{\arxivOnly}[1]{#1}
\newcommand{\ecaiOnly}[1]{}
\makeatletter
\def\blfootnote{\xdef\@thefnmark{}\@footnotetext}
\makeatother

\usepackage{fancyhdr}

\begin{document}

\title{Learning to Continually Learn}

\author{Shawn Beaulieu\institute{University of Vermont, USA, email: \{shawn.beaulieu, lapo.frati, ncheney\} @uvm.edu}
\and Lapo Frati$^1$ 
\and Thomas Miconi$^2$
\and Joel Lehman$^2$
\\
\and Kenneth O. Stanley\institute{Uber AI Labs, USA, email: \{tmiconi, joel.lehman, kstanley\} @uber.com}
\and Jeff Clune*$^{2,}$\institute{OpenAI, USA, email: jeffclune@OpenAI.com. Current affiliation (work done at Uber AI Labs).}
\and Nick Cheney*$^1$  
}

\maketitle
\bibliographystyle{ecai}

\thispagestyle{fancy}
\pagenumbering{gobble}
\renewcommand{\headsep}{18mm}
\renewcommand{\topmargin}{-18mm}
\lhead{Published as a conference paper at ECAI 2020}
\renewcommand{\textheight}{668px}

\begin{abstract}
Continual lifelong learning requires an agent or model to learn many sequentially ordered tasks, building on previous knowledge without catastrophically forgetting it. Much work has gone towards preventing the default tendency of machine learning models to catastrophically forget, yet virtually all such work involves manually-designed solutions to the problem. We instead advocate meta-learning a solution to catastrophic forgetting, allowing AI to learn to continually learn. Inspired by neuromodulatory processes in the brain, we propose A Neuromodulated Meta-Learning Algorithm (ANML). It differentiates through a sequential learning process to meta-learn an activation-gating function that enables context-dependent selective activation within a deep neural network.
Specifically, a neuromodulatory (NM) neural network gates the forward pass of another (otherwise normal) neural network called the prediction learning network (PLN). The NM network also thus indirectly controls selective plasticity (i.e. the backward pass of) the PLN. ANML enables continual learning without catastrophic forgetting at scale: it produces state-of-the-art continual learning performance, sequentially learning as many as 600 classes (over 9,000 SGD updates).

\end{abstract}

\section{Introduction}

Intelligent animals are typically able to solve new problems without corrupting hard-won knowledge of how to solve previously encountered problems. On the other hand, artificial neural networks have long been known to suffer from catastrophic forgetting (CF)~\cite{french1999catastrophic}, in which learning to solve new tasks rapidly degrades previously acquired capabilities. This problem is particularly acute for continual learning systems that encounter tasks sequentially. 
\blfootnote{*Co-senior authors} 
For that reason, deep learning's success stories have come in the non-sequential setting, with data sampled independently 
and identically distributed (i.i.d.) from the full dataset (aka interleaved, or shuffled, training).
A number of solutions to catastrophic forgetting have been developed, but they are typically \emph{manually designed}.

One such solution is to use replay methods \cite{schaul2015prioritized}, which alleviate forgetting by saving prior experiences and mixing them with newly encountered data to approximate interleaved (i.e. not sequential) training. However, the ability to interleave memories of all previously seen tasks between training examples on new tasks is expensive in both storage and computation, and does not scale to many tasks.

A second class of solutions attempt to limit the extent to which parameters can be modified in response to new data, which we refer to as \textit{selective plasticity}.  But such approaches typically involve manually designed heuristics that control such plasticity. 

In the extreme, all previously learned weights can be frozen and additional capacity can be added to the network to solve new tasks~\cite{rusu2016progressive}. 
Similarly, hand-designed modules within an overparameterized network can be frozen when they are used to solve a task~\cite{fernando2017pathnet}.
This approach also scales poorly as the number of tasks become large, and, by design,
old pathways cannot improve based on information from other tasks.
 
Rather than completely freezing its previously learned value, \textit{elastic weight consolidation} (EWC)~\cite{kirkpatrick2017overcoming}) alters the plasticity of a given parameter (indirectly, via regularization penalties) in proportion to 
the manually-selected criteria of Fisher information, which approximates
how important each parameter was for solving prior tasks. Those parameters deemed unimportant are made comparatively more susceptible to change, thus allowing the network to adapt to new problems while reducing corruption of existing knowledge. 
Other approaches similarly seek to modulate learning via task-specific synaptic importance~\cite{zenke2017continual}, via the top-down attention mechanism of contrastive excitation backpropagation~\cite{kolouri2019attention,zhang2018top},
by interleaving pseudo examples generated from the current weights to limit plasticity~\cite{li2017learning}, by employing L2 regularization on the weight changes between tasks~\cite{li2018explicit}, or by combining regularization and the Copy Weight with Reinit strategy~\cite{lomonaco2017core50, maltoni2019continuous}.

Another approach to mitigate CF is to directly incentivize the creation of maximally sparse or disjoint representations~\cite{french1991using, liu2018rotate}, with the goal of minimizing interference between activations. Such sparse representations indirectly affect which parameters get updated during backpropagation, thereby allowing the network to avoid forgetting. However, a perspective we advocate is that, when possible, we should not optimize for one thing (e.g. sparse representations) and hope doing so leads to another thing (in this case, reduced catastrophic forgetting): Instead, we should optimize directly for what we want (here, learning without forgetting). 

Rather than trying to manually implement a solution to CF or adding auxiliary losses we believe will alleviate CF, in this work we directly optimize to learn without forgetting. In other words, we harness meta-learning to allow the network to learn to continually learn. However, instead of vanilla meta-learning (e.g. MAML~\cite{finn2017model} with traditional neural networks), a contribution of this paper is introducing a new network architecture that improves the ability to learn to continually learn. Specifically,  we have one network, conditioned on the input, gate the activation of another network (i.e. context-dependent gating), resulting in \textit{selective activation} of that second network.
Gating functions for such conditional computation~\cite{bengio2013deep} have been learned via the REINFORCE algorithm~\cite{bengio2015conditional} and at scale via a standard backpropagation in a sparsely-gated mixture-of-experts~\cite{shazeer2017outrageously}, but these works focus on computational capacity and efficiency rather than catastrophic forgetting.

One prior work~\cite{MasseE10467} showed the promise of gating mechanisms for mitigating catastrophic forgetting, including high performance for up to 500 classes, although in an easier problem setting in which the learning algorithm is explicitly told which task the network is being trained or tested on, instead of having to learn that too. In realistic applications, such as a robot learning in the world, there will not be clear boundaries between tasks nor an oracle to provide task labels. Additionally, in this prior work, which neurons were gated was randomly chosen. Finally, the random gating was dependent on the task, not the data itself. In this work we remove the need to tell the network which task it is being trained or tested on, and we meta-learn the activation gatings, which provides the opportunity to substantially improve performance over random gating and enables the gatings to be dependent on the data itself. Finally, our learned masks are continuous and not binary (as in the prior work), further increasing the potential power of our method over binary, random gating.

While all of the above approaches are interesting and mitigate CF to some extent,
they nevertheless fall into the category of manual approaches to reduce CF.
That is, to solve a challenging problem they try to identify the required building blocks and how to combine them, and then let machine learning tune parameters within this complex assembly~\cite{clune2019ai}. However, a clear trend in machine learning is that manually designed solutions give way to entirely learned solutions, which ultimately perform much better, once sufficient compute and data are available. That has been true, for example, for learning features (e.g. in computer vision or speech recognition)~\cite{krizhevsky2012imagenet}, neural architectures~\cite{zoph2016neural,pham2018efficient}, and hyperparameters~\cite{maclaurin2015gradient}. 
An alternative to the manual path to AI that is in line with this trend of replacing hand-designed pipelines with learned solutions is to develop \emph{AI-generating algorithms}~\cite{clune2019ai}, which try to learn as much of the solution as possible, including learning the learning algorithms themselves via meta-learning. 

The meta-learning approach directly sets the ultimate goal (here: the ability to continually learn without forgetting) as a meta-loss and employs the power of machine learning to produce an AI algorithm well suited to it. An example is \textit{Model-Agnostic Meta-Learning} (MAML)~\cite{finn2017model}, which searches for an initial set of weights for a neural network that, when subjected to SGD, rapidly learns a new task. It does so by differentiating through many steps of inner-loop SGD learning to calculate the outer-loop gradient of how to improve the weight initialization. 

We are aware of only one prior work that optimizes to solve catastrophic forgetting via meta-learning. We were inspired by it, build off of it, compare to it, and adopt its experimental protocol. 
\textit{Online aware Meta-Learning} (OML)~\cite{javed2019meta} employs a MAML-style meta-learning algorithm to produce a representation (set of neural network layers) that, when frozen and used by additional, downstream neural network layers, minimizes catastrophic forgetting in those downstream layers. 
Excitingly, while sparsity in this representation layer was not explicitly encouraged, OML produced not only sparsity, but a better version of it. Explicitly encouraging sparsity yields many dead neurons (those that never fire across the dataset), whereas OML optimized representations had no dead neurons~\cite{javed2019meta}. 

This ability to search for what \emph{is} empirically effective, rather than what \emph{is believed} to be effective, distinguishes meta-learning approaches from manual-path approaches.
OML provided a substantial advance over the prior state of the art, enabling continual learning over 200 sequential classes. Because OML is in line with our vision of meta-learning solutions to hard machine learning problems, and because it performs so well, we seek to further improve upon it. 

Rather than meta-learning representations as in OML, we meta-learn a context-dependent gating function (a neural network, called the neuromodulatory network) that enables continual learning in another neural network (called the prediction network) (Fig.~\ref{fig:network_architecture}). The neuromodulatory network has the flexibility to explicitly turn on and off activations in subsets of the prediction network conditioned on the input. We call this mechanism selective activation. Selective activation in turn enables selective plasticity because the strength of backward gradients is a function of how active neurons were during the (modulated) forward pass, indirectly controlling which subset of the network will learn for each type of input. By leveraging meta-learning to optimize when and where to gate activations to maximize continual learning,
our approach explores the potential for solutions beyond hand-designed selective plasticity strategies. 

Our harnessing of selective activation via context-dependent gating and selective plasticity is inspired by neuromodulatory processes in the brain. These include inhibitory mechanisms that become active in response to specific environmental stimuli~\cite{MasseE10467} and the suppression of synaptic plasticity~\cite{hasselmo1995cholinergic} or activation~\cite{bear1986modulation} in the presence of neuromodulatory signals. There is prior work in machine learning that harnesses neuromodulation techniques~\cite{ellefsen2015neural, ishiguro2003neuromodulated, soltoggio2008evolutionary, soltoggio2018born, velez2017diffusion}, but in such work neuromodulation directly modulates learning rates, instead of the approach taken in this work of directly modulating activations and thus indirectly controlling learning. 
Modulating learning can ensure that information about one task can be localized to only the parts of the network relevant to that task. However, an insight in this work is that modulating learning alone is not enough, because there can still be interference between the tasks during the forward pass. For example, even if an agent's chess playing and bike riding networks are physically separated such that learning in one does not corrupt information in the other, neither task will be performed well if both are actively producing muscle outputs when performing either task. 
The insight behind why the approach in this work of directly modulating activations (and indirectly modulating learning) is superior is because it allows optimization to reduce interference in \emph{both} the forward and backward passes of the network.

\thispagestyle{ecai}
\renewcommand{\headsep}{0mm}
\renewcommand{\topmargin}{0mm}

\section{The Problem Formulation}
One goal of continual learning is to solve catastrophic forgetting, meaning learning new tasks without forgetting of already learned skills. More precisely, we want to be able to learn a large number of tasks $\mathcal{T}_{1..n}$ \emph{sequentially} from a common domain $\mathcal{T}$, but in such a way that after sequential learning average performance on all $\mathcal{T}_{1..n}$ tasks is high. 

Following~\cite{javed2019meta}, the experimental domain $\mathcal{T}$ is the Omniglot few-shot learning dataset.
Each task $\mathcal{T}_i$ is a class of characters (each with $k$ and $v$ unique training and validation instances, respectively). 

Before continuing, it is helpful to establish terminology for meta-learning, as it is complex. Because we are learning to learn, learning happens both on the outer loop and the inner loop. The goal is to have the outer loop take steps to make each round of inner-loop learning better. The phase in which the outer loop takes optimization steps to improve the learning ability of the inner loop is called \emph{meta-training}. Once meta-training has concluded, we then want to test the quality of the inner-loop learner, a phase called \emph{meta-testing}. During meta-training, within each inner loop, something must be learned, which is called \emph{meta-training training}. For example, the inner-loop agent might be a normal neural network learning via SGD to classify MNIST examples. During meta-training training, it is shown multiple MNIST samples (and given the correct label) and SGD steps are applied to improve accuracy. After each inner-loop iteration of meta-training training, the trained agent must be evaluated, which we call \emph{meta-training testing}. This meta-train testing error is the \emph{meta-training loss} to be minimized. After meta-training completes, we switch to meta-testing, wherein we need to evaluate how well the meta-learned learner performs. Again, we need a training and testing phase, which are called \emph{meta-test training} and \emph{meta-test testing}, respectively. We have found this MTT (for meta-train/test training/testing) terminology essential for clear communication and advocate the meta-learning community adopt it. 

A naive application of meta-learning to this problem would involve performing the entire learning process at each iteration of the inner loop: i.e. learn a whole sequence of $n \times k$ characters, evaluate the network's validation performance on $n \times v$ validation instances (the meta-training loss), backpropagate the gradient of this validation error all the way
through the learning process to the initial weights, perform one step of gradient descent on these initial weights, and repeat. However, when $n$ is high (e.g. 600), this approach is unfeasible due to limitations with modern algorithms (e.g. unstable gradients) and hardware (e.g. memory). 

The OML framework \cite{javed2019meta} introduced an elegant solution to this problem by setting the meta-loss to be an approximation of this true, desired meta-loss. The idea is to measure after each new class whether (1) it was learned and (2) whether old knowledge was lost. After each newly encountered class ($k$ instances), the meta-loss is calculated as the error on the newly learned class and the error on a random sample of characters from all meta-train training classes (called the \emph{remember set}). Ideally this remember set would only include previously seen meta-train training set images, but we follow the choice of \cite{javed2019meta} in sampling from all meta-training classes).

The remember set is used \emph{in meta-training only}. 

As a result, the network is meta-trained to learn a potentially large number of different classes without catastrophic forgetting, even though each inner-loop only involves learning $k$ instances of one character class, which makes the process computationally tractable. We use the same meta-learning procedure for training our novel architecture. 

The experiments employ the Omniglot few-shot learning dataset, which has 1,623 character classes~\cite{lake2015human}.  660 classes are held out for meta-testing. The remaining 963 are for meta-training.
For meta-train training, $k=20$, i.e. 20 labeled instances of each character are provided. The remember set randomly samples from these same 20 instances per class.  
In meta-testing (described below), 15 instances of each character are used for meta-test training, while the remaining 5 instances are held out for meta-test testing.  

\section{A Neuromodulated Meta-Learning Algorithm (ANML)}

While meta-learning provides the opportunity to learn the solution to a problem instead of manually designing it, we still must decide on the search space, including which \emph{materials} the neural networks are made of~\cite{clune2019ai}. To attempt to solve catastrophic forgetting, here we propose learning a neuromodulatory network, which is a context-dependent function that gates the forward pass of another (otherwise normal) neural network (Fig.~\ref{fig:network_architecture}). This setup allows different subnetworks within the normal network to be used for, and learn from, different types of tasks. It also allows the model to interpolate between tasks it has not previously encountered.  
By meta-learning, we automate, rather than presuppose, the contexts and locations within the network that should be active at any given time, and to what degree. 
We call this approach A Neuromodulated Meta-Learning algorithm (ANML).

\subsection{ANML Architecture}
\label{sec:ANML-OML-Architecture}

OML~\cite{javed2019meta} divides a single deep neural network into two distinct components. The first 6 layers, which are convolutional, have meta-learned parameters that are frozen (not changed) during the inner loop. This subnetwork is called the \emph{representation learning network} (RLN). The final 2 layers, which are fully connected, have parameters that meta-learn their initial weights in the outer-loop, and then are updated during each inner-loop via SGD. See Javed and White~\cite{javed2019meta} for details.  This subnetwork is called the \textit{prediction learning network} (PLN).  

\begin{figure}[t]
\includegraphics[width=\linewidth]{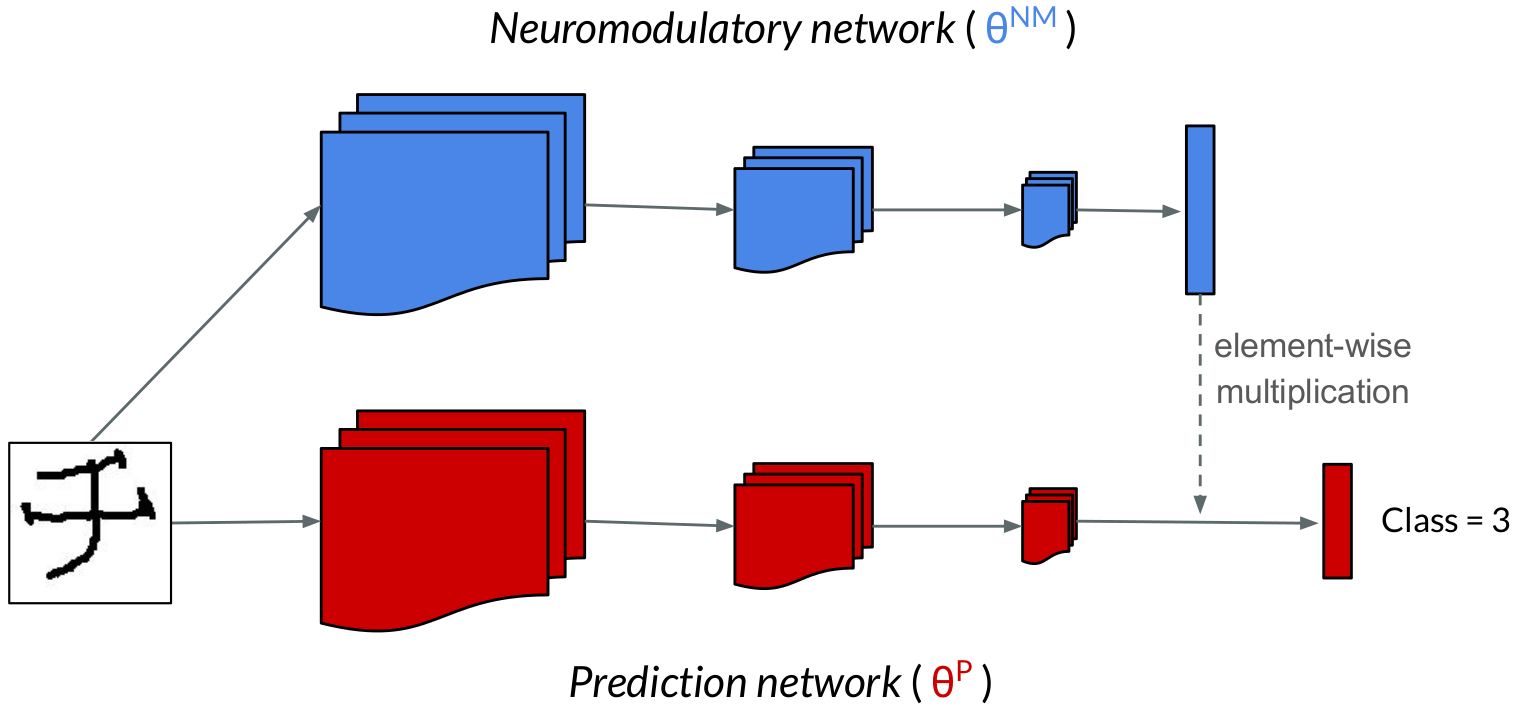}
\caption{
The architecture for \textit{A Neuromodulated Meta-Learning algorithm} (ANML).  The prediction network (red) is a normal neural network updated in the inner loop via SGD (or similar). The neuromodulatory network (blue) produces an element-wise gating of the prediction network's forward-pass activations, enabling selective activation (i.e. conditional computation) and indirectly enabling selective plasticity by affecting the gradient updates of the prediction network. The initial weights (at the start of each inner loop) of both the neuromodulatory and prediction network are meta-learned in the outer-loop of optimization. The weights of the neuromodulatory network are not updated in the inner loop, but those of the prediction network are. The example image is from the Omniglot dataset, which is the experimental domain for the experiments in this paper.}

\label{fig:network_architecture}
\end{figure}

ANML takes a different approach, with two parallel neural networks: a neuromodulatory (NM) network and prediction network (Fig.~\ref{fig:network_architecture}). The weights for both are meta-learned in the outer loop. 
The weights of the neuromodulatory network are not updated in the inner loop, but those of some, but not all, of the prediction network are (which prediction network weights are updated differs for meta-training and meta-testing, as described below). 
To keep overall architecture and parameter sizes similar to~\cite{javed2019meta}, each of these 
networks has 3 convolutional layers (each followed by a batchnorm layer~\cite{ioffe2015batch}) and one fully connected layer.
The final layer of the neuromodulatory network is of the same size as the input to the final layer of the prediction network (i.e.\ the flattened latent representation output by the final convolutional layer in the prediction network). The neuromodulatory output is used to gate the latent representation of the prediction network (via element-wise multiplication) during a forward pass. 
All activation functions are ReLUs except the gating multiplier is restricted to the range $[0,1]$ via a sigmoid, meaning that in this work it can only suppress activations of the prediction network (instead of negate or amplify them).  

\subsection{Meta-Training Learning Procedure}

The ANML algorithm (Algorithm~\ref{alg:ANML})\footnote{Code available at \url{github.com/uvm-neurobotics-lab/ANML}} consists of an inner loop nested inside an outer loop of optimization.  Within each inner loop, a copy $\theta^P_0$ of the prediction network weight initializations $\theta^P$ 
is trained for 20 SGD iterations -- producing $\theta^P_{1}$...$\theta^P_{20}$ -- on the training set of a single Omniglot meta-training class $\mathcal{T}_n$. 

During each of these 20 forward passes, the inputs to the prediction network's final layer are gated by neuromodulatory weights $\theta^{\text{NM}}$, enabling \textit{selective activation} by modifying the outputs of the prediction network during the forward pass.

During each backwards pass, the gating of the prediction network naturally has the effect of reducing the gradients flowing back towards a subset of its weights, thereby modifying the SGD update and resulting in \textit{selective plasticity}. The neuromodulatory weights $\theta^{\text{NM}}$ are not updated within the inner-loop.

Following the 20 sequential updates on this single Omniglot meta-training class, a meta-loss is calculated by making predictions using the final post-inner-loop-training weights ($\theta^P_{20}$) on all 20 images from the single Omniglot class just trained on plus a random sample of 64 character instances from the set of all meta-training classes (i.e. from the remember set).
This meta-loss function is referred to as the Online aware Meta-Learning (OML) objective~\cite{javed2019meta}, and incentivizes the meta-learning of networks that can learn a new tasks over many steps of SGD without forgetting how to solve previously learned tasks.

We then backpropagate this meta-loss back through the 20 steps of SGD updates (following the style of MAML~\cite{finn2017model}) to calculate the gradients of the initial prediction network weights $\theta^P$ and the neuromodulatory network weights $\theta^\text{NM}$, and then take an outer-loop gradient update step on each of them.  This outer-loop step is taken via Adam~\cite{kingma2014adam}, while gradient updates for all inner-loop updates are via SGD with a fixed learning rate.  This step constitutes the completion of the first outer-loop iteration. The next outer-loop iteration then begins by copying the prediction net weights and conducting inner-loop training starting with them on 20 instances from the next Omniglot meta-training class.  For the experiments in this paper, this process continues for 20,000 outer-loop meta-training iterations. 

Following OML~\cite{javed2019meta}, when a copy of the prediction network is made at the start of each meta-iteration, the weights in the final layer that lead into the output node for the single Omniglot class in the upcoming meta-train training trajectory are initialized randomly in that inner-loop copy (rather than being initialized with their meta-learned weight initialization). This ensures that the inner-loop learner has not already converged to a good solution to classifying images in the upcoming trajectory through its initial-weight-meta-learning alone, and will result in larger gradient steps during backpropagation -- both of which more closely match the situation that will be encountered during meta-testing, when each class encountered is new. Note that reinitializing the entire final layer would not be ideal, as it would not enable the network to make intelligent predictions about the remember set (which contains classes not trained on during this inner-loop).  This weight reinitialization procedure is not employed during meta-test-training.  

\subsection{Performance Evaluation in Meta-Testing}

\begin{algorithm}[t]
\caption{A Neuromodulated Meta-Learning algorithm (ANML)\label{alg:ANML}}
\begin{algorithmic}[1]
\Require $\mathcal{T}\gets$ trajectory of $N$ sequential meta-training tasks
\Require $\theta^{\text{NM}} \gets $ weights of the neuromodulatory network
\Require $\theta^{P} \gets $ weights of the prediction network
\Require $\alpha,\beta \gets $ learning-rate hyperparameters
\State initialize $\theta^{\text{NM}}$, $\theta^{P}$
    \For{$n=1,2, \ldots$} \Comment{meta-learning outer-loop} 
        \State $S_{traj} = \mathcal{T}_n$ \Comment{trajectory for inner-loop training}
        \State $S_{rem} \sim \mathcal{T}$ \Comment{sample instances from all tasks to remember}
        \State $\theta^{P}_{0} = \theta^{P}$ \Comment{create inner-loop copy of prediction net}
        \For{$i=1,2, \ldots,k$} \Comment{task-learning inner-loop}
            \State $\theta^{P}_{i} \gets \theta^{P}_{i\text{-}1} - \beta
            \nabla_{\theta^{P}_{i\text{-}1}}\mathcal{L}(\theta^{\text{NM}},\theta^{P}_{i\text{-}1},S_{traj}$) \Comment{SGD on $\theta^{P}_{i\text{-}1}$}
        \EndFor
        \State $\theta^{\text{\text{NM, P}}} \gets \theta^{\text{NM, P}}-\alpha \nabla_{\theta^{\text{NM, P}}}\mathcal{L}(\theta^{\text{NM}},\theta^{P}_{k},S_{traj},S_{rem})$
       
        \Comment{meta-update on $\theta^{\text{NM}}$, $\theta^{\text{P}}$ w.r.t. final inner-loop $\theta^{P}_k$}
    
\EndFor
\end{algorithmic}
\end{algorithm}

\begin{algorithm} [t]
\caption{Meta-Testing Evaluation Protocol\label{alg:Eval}} 
\begin{algorithmic}[1]
\Require $\mathcal{T} \gets$ trajectory of $N$ unseen sequential meta-testing tasks
\Require $\theta^{\text{NM}} \gets $ meta-learned weights of the neuromodulatory net
\Require $\theta^{P} \gets $ meta-learned weights of the prediction network
\Require $\beta \gets $ learning-rate hyperparameter

\State $S_{train} = [\ ] $ 
\For{$n=1,2, \ldots,N$}  
    \State $S_{traj} \sim \mathcal{T}_n$ \Comment{get next task training trajectory}
    \State $S_{train} = S_{train}+S_{traj}$ \Comment{add to meta-test train set}
     \For{$i=1,2, \ldots,k$}      
        \State $\theta^{P} \gets \theta^{P} - \beta \nabla_{\theta^{P}}\mathcal{L}(\theta^{\text{NM}},\theta^{P},S_{traj})$
        \Comment{SGD on $\theta^{P}$}

    \EndFor
    \EndFor
        \State record $\mathcal{L}(\theta^{\text{NM}},\theta^{P},S_{train}) $
        \Comment{eval final $\theta^{P}$on meta-test train set}
    \State $ S_{test} = \mathcal{T} - S_{train} $ \Comment{held-out meta-test test set}
        \State record $\mathcal{L}(\theta^{\text{NM}},\theta^{P},S_{test}) $
        \Comment{eval final $\theta^{P}$on meta-test test set}

\end{algorithmic}
\end{algorithm}

Following the completion of meta-training, in the meta-test phase (Algorithm~\ref{alg:Eval}) the resulting prediction and neuromodulatory networks are evaluated on their ability to learn many tasks while minimizing catastrophic forgetting.  

Starting from the meta-learned prediction network weight initialization $\theta^P$ and neuromodulatory network $\theta^{\text{NM}}$, the weights of the fully connected layer 

in the prediction network (and only those weights) are fine-tuned on meta-test classes. This idea of freezing the weights of all but the last few layers of the prediction network follows~\cite{javed2019meta}.  These weights are fine-tuned for $q$ (here, 15) instances of each meta-test class (here, the 600 Omniglot classes that were not used during meta-training). 
Unlike meta-training, in meta-test training, a new copy of the prediction network is not made for each new class. Thus, in our experiments, the final weights from meta-training are fine-tuned during meta-test training with 9,000 iterations of SGD.  The model therefore undergoes 8,985 fine-tuning updates since it last saw an instance from the first Omniglot meta-test training class.  Using these final weights ($\theta^P_{9000}, \theta^{\text{NM}}$), all 9,000 meta-test training instances are reevaluated to assess the meta-test training performance of the model (i.e.\ 
how well it can learn the training set without forgetting). Furthermore, the final model ($\theta^P_{9000}, \theta^{\text{NM}}$) is evaluated on five held-out instances from each of the 600 meta-test classes (the meta-test test set), which measures the model's ability to generalize what it has learned to novel instances of each class. In short, the meta-test training performance shows the ability to memorize without forgetting, and the meta-test test performance reveals the ability to conduct continual learning in a way that allows generalization, which is what we ultimately care most about.

Following OML's evaluation protocol~\cite{javed2019meta}, this procedure is repeated for various sequence lengths of meta-test classes (trajectories of 10, 50, 75, 100, 150, 200, 300, 400, 500, and 600 Omniglot classes) to show how the models scale to longer sequential task trajectories.  A hyperparameter search is performed for each sequence length to set the learning-rate $\beta$ for its inner-loop updates. 

\subsection{Baseline Controls}

To help analyze the effectiveness of ANML, we  compare it to a series of controls. All architectures in the control treatments share approximately the same numbers of total parameters as ANML ($\sim$6M). They do not include a neuromodulatory network, but instead 
have the OML architecture (described in Section~\ref{sec:ANML-OML-Architecture}). Empirically, we found that the batchnorm layers included in ANML were detrimental to the performance of OML, so they were omitted from these controls (consistently with~\cite{javed2019meta}).  

\textbf{Training from Scratch:} 
Consistently with traditional machine learning, this control involves no meta-training and simply randomly initializes a prediction network at the start of meta-test training time.  In this treatment the entire network can learn, rather than fine-tuning only the fully-connected layers, as the network does not include any learned features at the start of meta-test training.

\textbf{Pretraining and Transfer:} 
The previous control is perhaps unfair because ANML gets to learn from the meta-training set before being evaluated on the meta-test set. The Pretraining and Transfer control tries to address this inequity. Consistently
with traditional transfer learning for deep neural networks~\cite{yosinski2014transferable}, this control pretrains i.i.d. on the meta-training training image set. The number of images seen during pretraining is set to be equivalent to the numbers of instances seen during meta-train training and meta-train testing (1.68M image evals). 

The weights learned during this pretraining phase are transferred to the meta-test phase, where the fully connected layers are fine-tuned on the meta-test training trajectory.

\textbf{Online aware Meta-Learning (OML):} 
OML~\cite{javed2019meta} represents the current state-of-the-art on this problem domain.  ANML has the same meta-learning procedures as it, with the exception that OML does not include a neuromodulatory network ($\theta^{\text{NM}}$) for selective activation, and instead meta-learns the weights of the convolutional layers only, which they call the Representation Learning Network (RLN)~\cite{javed2019meta}). 
The RLN weights in OML are frozen at meta-test time and within each inner-loop of meta-train training: they are modified only via the outer-loop updates during meta-training. Similarly, ANML freezes the convolutional weights and batch norm weights of the prediction network during meta-testing, only updating the weights of the fully connected layer of the prediction network during this meta-test phase. However, in contrast to OML, in ANML no weights in the prediction network are frozen during meta-train training. Additionally, as with the parameters in OML that are only meta-learned (the RLN), the weights of the neuromodulatory network in ANML are frozen during both meta-testing and meta-train training (they are only updated via outer loop steps during meta-training).

The meta-test procedure for the original OML algorithm includes fine-tuning both fully-connected layers in the prediction network at meta-test time, while freezing the 6 convolutional layers.  However, fine-tuning two layers (vs.\ one) leads to approximately 50\% more fine-tuned parameters ($\sim$3M) compared to the ANML architecture (which fine-tunes $\sim$2M parameters).  To counterbalance this, we also examine a treatment that fine-tunes only the final layer of OML, which results in about 50\% fewer fine-tuned parameters ($\sim$1M) than in ANML.

We call this control treatment \textit{OML with One-Layer Fine-Tuned} (OML-OLFT). The OML-OLFT meta-testing procedure often (but not always) results in improved performance compared to the original OML, but never results in OML-OLFT outperforming ANML. Because it is the published previous method,  we refer to the original OML formulation when making statistical comparisons to OML in the text (unless otherwise noted), but we show both in plots.    

\textbf{Interleaved-Training Oracles:}
To understand the limits of this problem domain, the 
\textquote{Oracle} version of any of the above algorithms keep their same meta-training procedure, but at meta-test time are presented with an i.i.d. sampling of all the classes from the meta-test training set.  These treatments thus examine the ability of an algorithm to learn without the challenge of catastrophic forgetting from sequential/continual learning, and provide an upper bound on the performance on a model with respect to avoiding catastrophic forgetting.

\section{Results}
\subsection{Continual Learning at Scale}

To our knowledge, the longest continual learning trajectories reported to date that exhibit robustness to catastrophic forgetting involve learning 200 tasks sequentially and come from OML~\cite{javed2019meta}.\footnote{Sequences up to 500 have been tried, but only when providing a knowledge of which unique task is being solved directly to the network instead of requiring it to learn that too~\cite{MasseE10467}.} Here we push that further, testing whether 600 sequential tasks can be learned -- seeking sequential learning across up to 9,000 SGD updates without catastrophic forgetting.  

For each treatment, we meta-train 10 independent models on the same set of 963 meta-training classes.  At the \emph{end} of the meta-test \emph{training} trajectory, we re-evaluate the trained models on all of the meta-test \emph{training} data seen over the trajectory to see how well they resist catastrophic forgetting. ANML significantly outperforms OML for meta-test training trajectories at all lengths tested (all $p\leq 1.26\times10^{-8}$; all p-values computed using the Mann-Whitney U test~\cite{mann1947test}).

\begin{figure}[t]
\includegraphics[width=\linewidth]{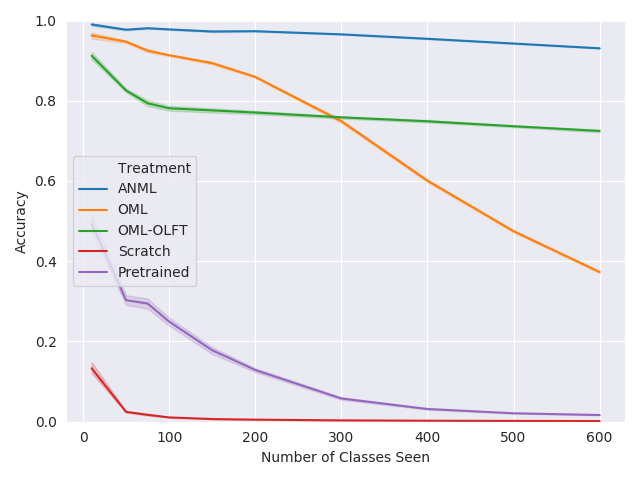}
\caption{
Meta-test \emph{training} classification accuracy.  The $x$-axis shows the number of sequential tasks/classes in the meta-test training trajectory.  Accuracy is calculated with the final prediction network parameters after training sequentially on that full meta-test training trajectory, and on all instances in the that trajectory (i.e. the meta-test training set).
}
\label{fig:meta-test training}
\end{figure}

Consistent with the expectation that standard neural networks suffer from catastrophic forgetting, both pretraining-then-transferring learned representations, and training randomly initialized networks from scratch have significantly worse meta-test training accuracy than either OML and ANML for all class sequence lengths tested (all $p\leq 6.023\times10^{-20}$), resulting in very low classification accuracy (Fig.~\ref{fig:meta-test training}; $<$ 3\% on trajectories of $\geq 50$ sequential classes for \emph{Scratch}, and $<$ 3\% on trajectories of $\geq400$ sequential classes for \emph{Pretrain})

Interestingly, the OML-OLFT treatment, which fine-tunes fewer parameters than standard OML, has higher meta-test training accuracy than OML on trajectories of 300 or more classes, but performs significantly worse on this metric for trajectories of up to and including 200 classes. All $p\leq 1.87\times10^{-13}$ up to 200 classes for OML relative to OML-OLFT; and all $p \leq 1.33\times10^{-29}$ for trajectories equal to 400 or more for OML-OLFT relative to OML.

\begin{figure}[!t]
\includegraphics[width=\linewidth]{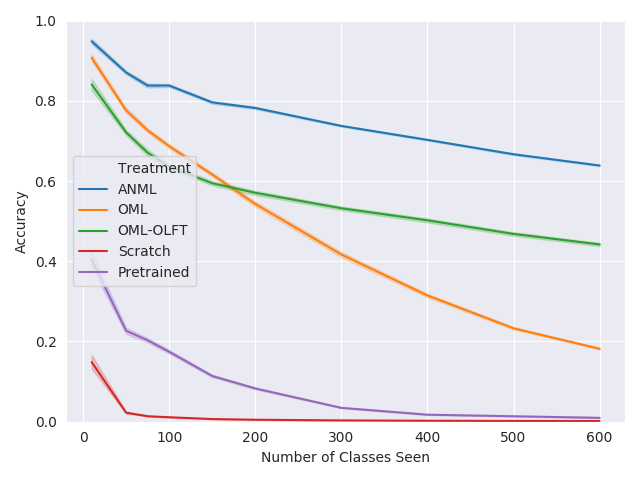}
\caption{
Meta-test \emph{testing} classification accuracy.  The $x$-axis shows the number of sequential tasks/classes in the meta-test training trajectory.  Accuracy is calculated with the final prediction network parameters after training sequentially on that full meta-test training trajectory, and the evaluation is on held-out (i.e. test) instances of the meta-test classes. Thus, these meta-test test instances were not seen during meta-training or meta-test training.  For all trajectory lengths tested, ANML significantly outperforms OML, the pretrained-and-transfer networks, and models trained from scratch.
}
\label{fig:meta-test testing}

\includegraphics[width=\linewidth]{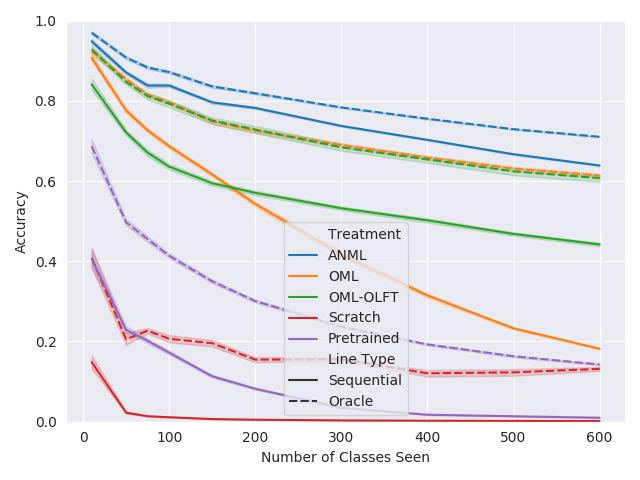}
\caption{
Meta-test \emph{testing} classification accuracy vs. oracles. The $x$-axis shows the number of sequential tasks/classes in the meta-test training trajectory.  Accuracy is calculated with the final prediction network parameters after training on that full meta-test training set i.i.d. for all \textit{oracle} treatments (and after sequential training for the ANML treatment).  Accuracy is reported for held-out instances of the meta-test classes not seen during meta-test training or meta-training.  ANML outperforms the i.i.d.-trained oracle versions of all other treatments.  
}
\label{fig:meta-test testing-oracle}
\end{figure}

\begin{figure*}[!ht]
\begin{center}
\includegraphics[width=\linewidth]{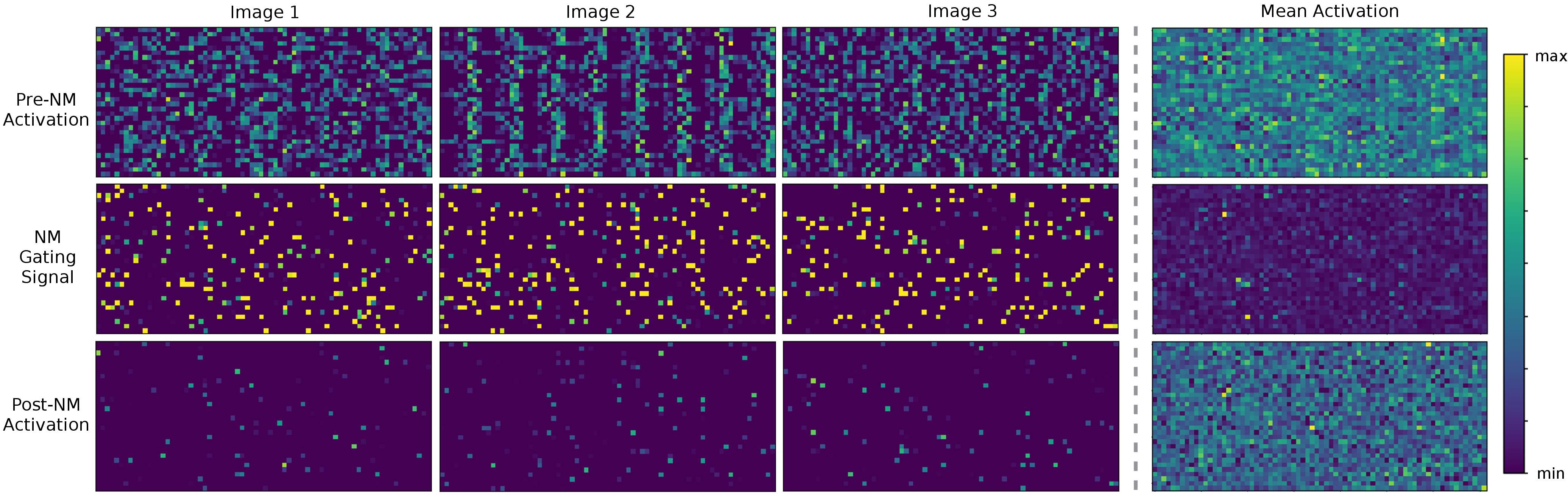}
\caption{The sparsity of activations before (top row) and after (bottom row) the neuromodulatory gating signal (middle row) has been applied, shown for three random inputs from the meta-test test set, and the mean across all images in the meta-test test set.  Colorbars for subfigures are individually normalized to better show min (blue) and max (yellow) activations.  Note that post-NM activations are sparse for each individual image, but near-uniformly distributed on average, revealing that NM helps to create sparse, orthogonal representations, and efficiently uses all of its compute resources instead of creating wasteful ``dead neurons.''
}
\label{fig:activations-pre-post-nm}
\end{center}
\end{figure*}

We now examine meta-test \emph{testing}, which involves the more difficult challenge of learning (without forgetting) in a way that \emph{generalizes} to never-before-seen instances of classes learned during meta-test training. 
For each of the 10 meta-training trials of each algorithm, we evaluate the meta-trained model on 10 independent meta-test training trajectories, each with a random set and order of tasks drawn from the 660 meta-test classes that were held out from the meta-training dataset.

For trajectories of any length, ANML has significantly better meta-test \emph{test} accuracy than any other treatment including OML (Fig.~\ref{fig:meta-test testing}, $p\leq 2.58\times10^{-12}$).
After meta-test training on 600 classes, ANML is able to correctly classify a mean of 63.8\% of held-out meta-test test instances, compared to the 18.2\% for OML and 44.2\% for OML-OLFT.  

Furthermore, ANML significantly outperforms chance ($p<7.96\times10^{-6}$) on 99.3\% of classes over the 600-task sequence, including many classes that it has not seen for hundreds of tasks (thousands of SGD updates), demonstrating learning without forgetting catastrophically\ecaiOnly{\footnote{\label{note1}Figure available in supplementary material of arXiv manuscript}}\arxivOnly{ (Fig.~\ref{fig:errorDistribution})}.

At meta-test time, the OML-OLFT treatment is significantly inferior to OML for trajectories of up to 150 classes ($p\leq 6.23\times10^{-8}$), but is better able to scale to longer trajectories, outperforming OML when sequentially learning 300 or more classes ($p \leq 4.69\times10^{-12}$ for $\geq 300$ classes).
While OML-OLFT fine-tunes fewer parameters during meta-test training, additional experiments explore fine-tuning of a greater number of parameters for both the OML and ANML networks during meta-test training find that ANML maintains a greater proportion of its performance when some or all additional layers are also fine-tuned\ecaiOnly{\footnote{Figure available in supplementary material of arXiv manuscript}}\arxivOnly{ (Fig.~\ref{fig:robustness})}.

Given that ANML significantly reduces catastrophic forgetting, a natural question to ask is what the upper-bound on performance is when catastrophic forgetting isn't a problem -- that is, when data is presented non-sequentially. The ideal version of any model, or \textquote{oracle}, can be approximated by training it in an interleaved fashion; replacing the sequential trajectory of correlated images with i.i.d. samples from the set of all images in this sequence. 
As expected, the i.i.d. oracle version of each algorithm significantly outperforms its sequentially-trained counterpart (Fig.~\ref{fig:meta-test testing-oracle}, $p\leq 1.27\times10^{-34}$ for all treatment vs. oracle pairs for sequences of length 600).  However, the sequentially-trained version of ANML significantly outperforms even the oracle versions of OML and the other algorithms at 600 classes (all $p\leq 1.93\times10^{-23}$).  The only treatment that outperforms ANML is the ANML-Oracle (for 600 classes, ANML-Oracle: 71\%, ANML 63.8\%, $p= 1.27\times10^{-34}$).
The finding that ANML produces models that perform better than traditional approaches even when catastrophic forgetting is not an issue hints at the promise of the ANML approach to also provide performance improvements to traditional i.i.d. training tasks in addition to the continual learning setting, which is an interesting area for future work.  

Another useful metric is the relative \emph{drop} in performance from the i.i.d. oracle to the sequentially-trained version of each algorithm.  Presuming that the primary difference between these treatments is the presence of catastrophic forgetting, the smaller the relative performance drop between them, the more the sequentially trained version of the algorithm has avoided catastrophic forgetting.  Of the relative drops in performance on trajectories of 600 classes for the sequentially learned versions of training-from-scratch (99\% relative drop in performance from i.i.d. to sequential learning), pretraining-and-transferring (99\%), OML (70.32\%), OML-OLFT (27.2\%), and ANML (10\%), ANML shows the smallest relative drop in performance between the i.i.d. and sequentially-trained version (all $p\leq 3.11\times10^{-24}$ relative to ANML), further supporting the benefits of ANML over the other algorithms for continual learning tasks. It is remarkable to see such a small performance drop (only 10\%) for ANML between the i.i.d. and sequential version of the task, meaning that ANML is solving most of the catastrophic forgetting problem on this challenging test across 600 sequential tasks.

Traditionally, continual learning in neural networks involves seeing examples in a single continuous stream -- therefore, the oracle treatments described above see each meta-test training image \emph{exactly once}. That is one reason why the performance is so low for standard deep learning techniques, such as training from Scratch and Pretraining \& Transferring, even with i.i.d. sampling (Fig.~\ref{fig:meta-test testing-oracle}). With additional passes through the data, performance increases across the board. 
After 20 epochs of i.i.d. training on the 600-class meta-test training dataset, training from Scratch gets 61.8\% accuracy on the meta-test test set, while the Pretrain \& Transfer control results in 48.66\% accuracy. 
While these accuracy levels may seem low, recall that the Omniglot meta-test training set here contains 600 classes, each with \emph{just 15 training instances}.  By comparison, the relatively high training accuracies (e.g. 99.77\% at 500 epochs of training~\cite{ciregan2012multi}) on handwritten digits in the MNIST dataset~\cite{lecun1998gradient} result from a dataset with just 10 classes, each with 6,000 training instances. Because the number of training instances per class is low, the Scratch and Pretrain controls overfit. Evidence of overfitting is that, while their test accuracy is low, their performance on the meta-test \emph{training} set is near-perfect (for Scratch accuracy is 99.3\%; for Pretrain \& Transfer it is 98.9\%). 

In this context, it is remarkable how well ANML performs. With just one pass through the data, and with the data ordered sequentially instead of i.i.d., the accuracy of ANML (63.8\%) is still higher than the Scratch and Pretrain \& Transfer 20-epoch, i.i.d. controls. The performance of ANML further increases when it has the benefit of training \emph{i.i.d} for one epoch  (Fig.~\ref{fig:meta-test testing-oracle}; 71\%). With 20 epochs, the accuracy increases further, to 75.37\% for ANML.
ANML continues to significantly outperform all other treatments with more epochs of training\ecaiOnly{\footnote{Figure available in supplementary material of arXiv manuscript}} (\arxivOnly{Fig.~\ref{fig:multiple-epoch-oracles}, }$p=3.38\times10^{-12}$). It is an open, interesting, important research question to figure out exactly why ANML outperforms the controls even in the multi-epoch, i.i.d. setting. 

\subsection{The Meta-Learned Representations}
\label{sec:representations}

\begin{figure*}[t]
\includegraphics[width=0.33\linewidth]{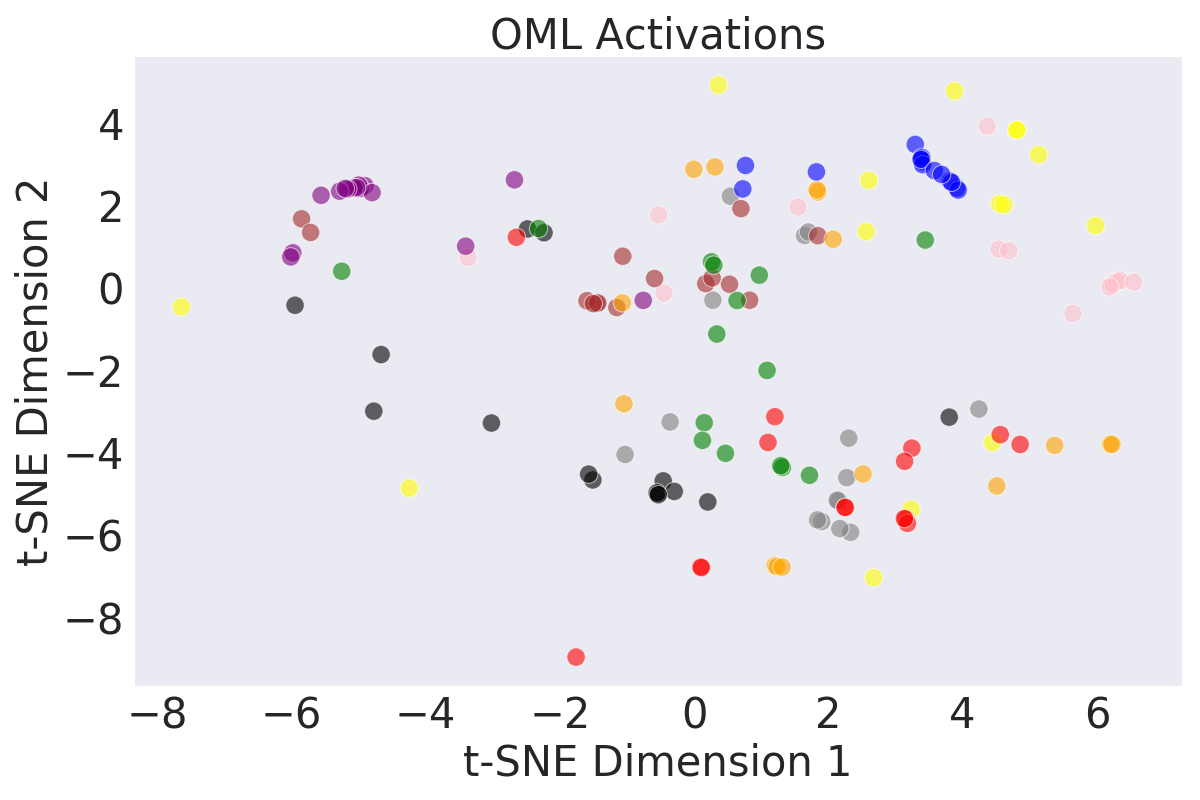}
\includegraphics[width=0.33\linewidth]{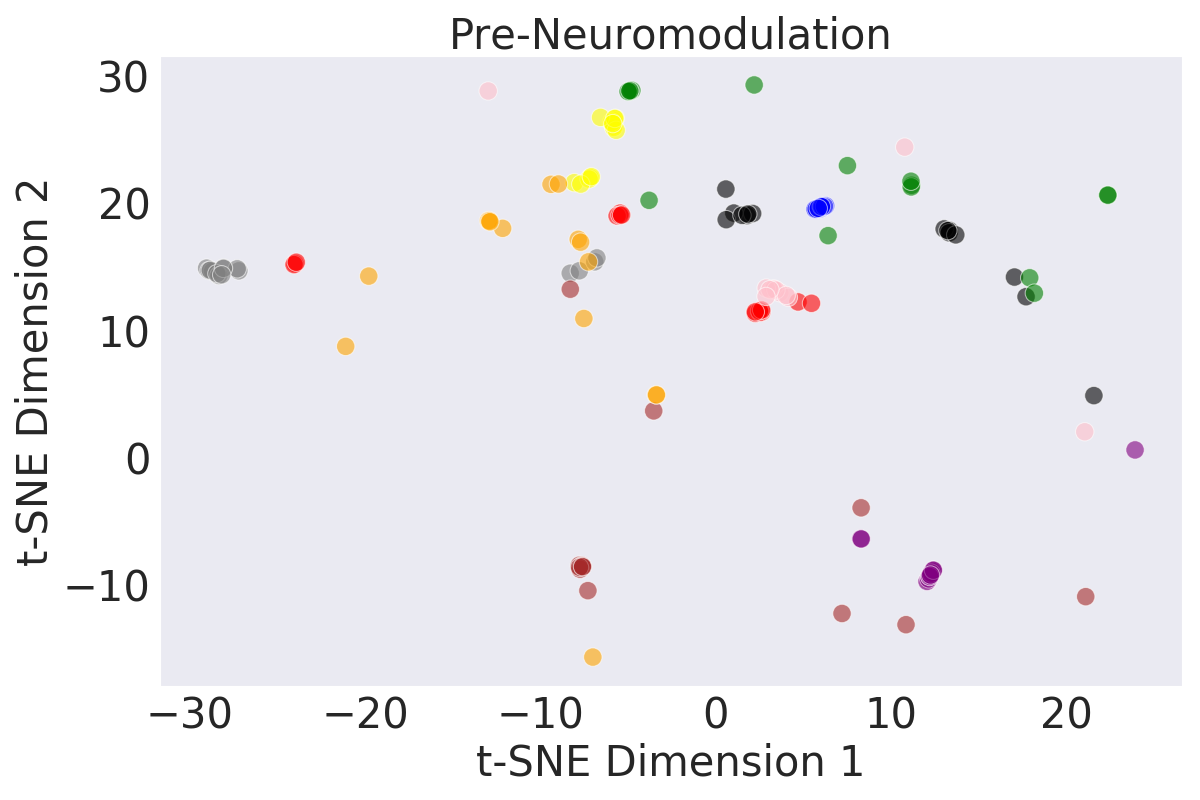}
\includegraphics[width=0.33\linewidth]{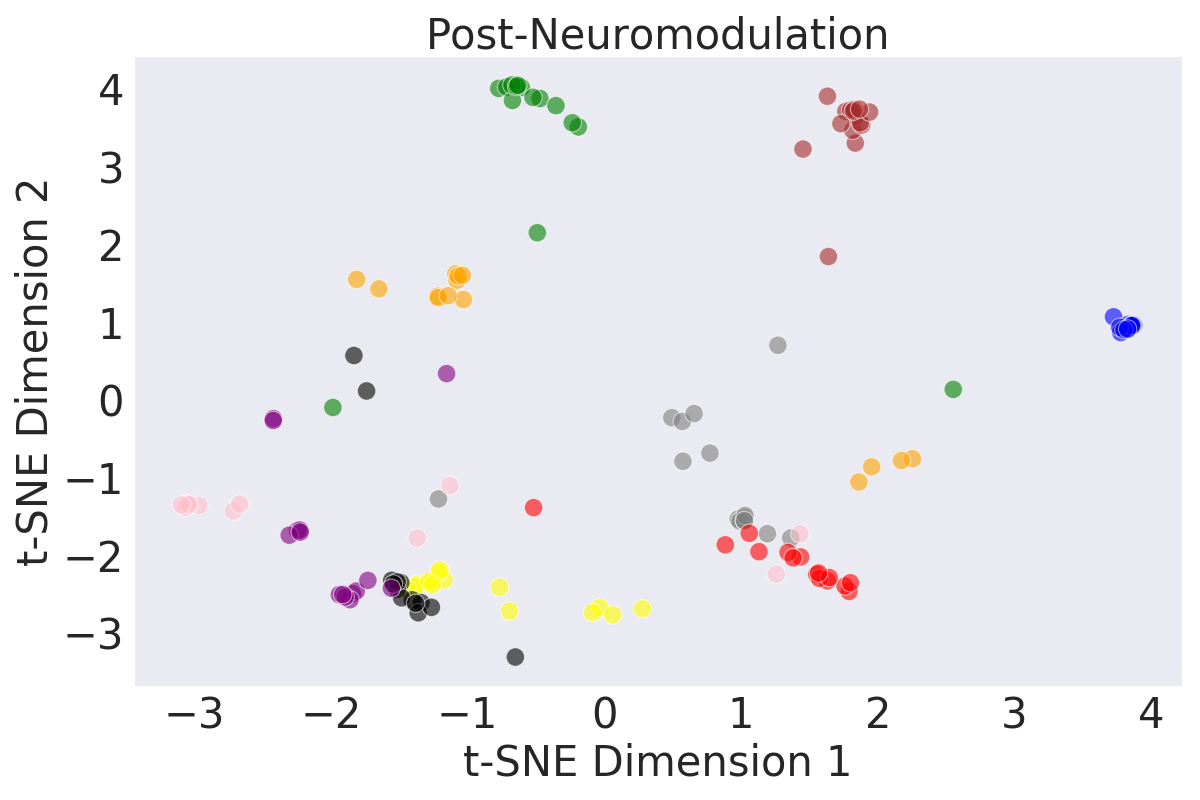}
\caption{2D t-SNE projections of the latent 
representations for 10 randomly selected meta-test training classes when the network saw them for the first time (before any labels are provided). \textit{Left:} OML representations. \textit{Middle:} PLN representations before being gated by the neuromodulatory network. \textit{Right:} PLN representations after being gated by the neuromodulatory network. Qualitatively, the post-neuromodulatory activations provide more well-separated clusters. KNN classification accuracy quantitatively shows the improved accuracy that results from such separation (see text). 
}
\label{fig:tsne-pre-post-nm}
\end{figure*}

Prior work has shown that encouraging sparse representations (e.g. via an auxiliary loss) alleviates catastrophic forgetting~\cite{french1991using, french1999catastrophic, liu2018rotate}. One of the key findings of OML~\cite{javed2019meta} was that, despite not explicitly incentivizing representations to be sparse, when meta-learning directly for the reduction of catastrophic forgetting, the system learned on its own to produce sparse representations.  Here we investigate whether sparse representations also arise in ANML without an explicit incentive for sparsity.

To examine the output of the neuromodulatory network, and how its gating affects the sparsity of the prediction network's representation, we analyze the activations of both networks in response to the images in the meta-test test set. The prediction network's representation layer \emph{prior} to neuromodulation has a mean of 52.77\% of neurons active (defined as activations $>0.01$). \emph{After} neuromodulation, the proportion of neurons in this layer that are active drops significantly to 5.9\% on average ($p<10^{-6}$).
By comparison, the sparsity of the representation in OML to these same images is 3.89\%.

The average number of neurons active in the last layer of the NM network (which does the gating) is 56.9\%.
Finally, both OML and ANML have 0\% ``dead neurons'' at the representation layer during meta-test training time. This means that every neuron is active for at least one class in the meta-test training set. In contrast, directly encouraging the production of sparse representations creates dead neurons, which are wasteful, at much higher frequencies \cite{javed2019meta}, providing more evidence that it is better to optimize directly for what we want (learning without forgetting), rather than optimizing for one thing (sparsity) and hoping we get an optimal version of something else. 

For a randomly chosen meta-training trial and for three different example input images (randomly chosen from the meta-test test set), the activations of the neuromodulatory network and the pre- and post-neuromodulatory-gating activations of the prediction network are shown in Fig.~\ref{fig:activations-pre-post-nm}. This visualization, in combination with the average activation rate reported above, show that ANML has indeed learned to produce sparse representations simply through a incentive to directly maximize continual learning, as OML did~\cite{javed2019meta}. Furthermore, that ANML is more successful at avoiding catastrophic forgetting than OML, yet does so with a less sparse representation, suggests that having sparse representations alone are insufficient.  ANML is able to meta-learn a combination of sparse representations and selective plasticity that together are especially important for continual learning.

We hypothesize that the neuromodulation networks learns to recognize different types of images, and creates different selective activation masks for them. That would, in turn, enable \emph{different parts of the prediction learning network} to store information about \emph{different types} of images, reducing catastrophic forgetting (recall that selective activation leads to selective plasticity). A prediction of this hypothesis is that the outputs of the neuromodulatory network should be different for different types of images (i.e.\ that it should be possible to cluster semantically similar images by their neuromodulatory outputs). 

To test this prediction, we use the K-Nearest-Neighbor (KNN) algorithm~\cite{fix1951discriminatory} to predict the class label of images from the meta-test test set, using the Euclidean distance in the
space of NM activations (labels come from the meta-test training set).
The following results are with $K=5$, but the results are qualitatively unchanged across $1 \leq K \leq 20$).
 
The result show that the NM outputs enable an accuracy of 70.9\%, which is significantly higher than chance. The accuracy of a neural network with the same architecture as the NM network, but with random weights is 24.3\% ($p=2.58\times10^{-31}$). This result confirms that the NM network has meta-learned the ability to tell apart types of images, including classes it has never seen before, enabling it to create different selective activation and plasticity for different types of images. Interestingly, the separability the neuromodulatory network creates also makes it easier for the prediction network to learn to solve the task (because the classes are better separated). 

Another test of the value of the NM network is if it \emph{improves} the representations in the PLN. To test that, we compare KNN accuracy based on the PLN activations before and after they are gated by the NM network. The KNN classification accuracy post-neuromodulation is 81.1\%, which significantly outperforms the 57\% classification accuracy of pre-neuromodulation activations  ($p=7.87\times10^{-12}$). 

The improvement to class separability post-neuromodulation can also be seen visually in 2-D t-SNE~\cite{maaten2008visualizing} projections of the representation layer activations for 15 instances each of 10 meta-test classes when the network (resulting from a single randomly chosen meta-training trial) saw these instances \emph{for the first time} during meta-test-training (Fig.~\ref{fig:tsne-pre-post-nm}). 

The meta-learned representations in the PLN already do a good job of separating the never-seen-before classes, which makes it easier to learn to perform the task well. The meta-learned NM network further increases that separability via selective activation. That, in turn, creates selective plasticity, causing information about each class to be stored in different parts of the network, reducing catastrophic forgetting. 

\section{DISCUSSION AND FUTURE WORK}

The results above demonstrate ANML's impressive ability to continually learn, outperforming the current state-of-the-art approach of OML on this domain~\cite{javed2019meta}. 
The results show the promise of meta-learning selective activations (and by extension, selective plasticity) to help reduce catastrophic forgetting and enable continual, lifelong learning in deep neural networks.
In future work it would be interesting to study the extent to which ANML and meta-learning in general improve other aspects of continual learning, such as forward transfer (being better on future tasks due to prior experience) and backwards transfer (getting better on previously learned tasks when learning new tasks).
Sequentially-trained ANML even outperforms the i.i.d. trained OML-Oracle, suggesting that meta-learned neuromodulation and its resulting selective activations may be powerful for problems beyond reducing catastrophic forgetting,
and may represent a powerful architecture more generally.

ANML's absolute performance of 63.8\% accuracy on held-out images from 600 Omniglot classes would not be impressive in an i.i.d., multi-epoch, large-training-dataset context.  However, it is impressive considering (1) that the data are presented sequentially, including over nearly 9,000 SGD updates, (2) that each image is seen just one time, and (3) that the number of training instances per class is low (at just 15). ANML's achievement is clearer when comparing to the performance of traditional deep learning methods (the Scratch or Pretrain \& Transfer controls) in this difficult task setting. 

There are many directions for future work. Initially, while there is only a
10\% drop in performance owing to catastrophic forgetting (i.e. between the ANML-Oracle and ANML), that still leaves room for improvement. There is also work to be done in analyzing the implementation choices made during this study, which could further improve performance. For example, the choice to gate selective activations in just one layer of the prediction network was a simplifying assumption made in this initial exploration, but the methodological approach can easily be extended to modulate any and all layers within the prediction network. Additionally, the technique could be made more fine-grained.
Indeed, during the development of ANML, a version of neuromodulation was created where every synapse in the prediction network was gated by the NM network. This version was competitive with other versions that were tested, but required too many parameters to be a viable alternative given our current computational resources. To avoid this explosion in the parameter count, one could harness indirect encoding, as in HyperNEAT~\cite{stanley2009hypercube}, where a small network can create a geometric pattern (here, of gating) that can be applied to a larger network. Interestingly, this approach would also enable the NM network to be able to be applied to prediction networks of different sizes at meta-test time.   

So far ANML has been demonstrated in a situation where each class/task is simple (learning one Omniglot character type). New research is needed to see how well ANML and ANML-inspired approaches scale to much more complex tasks. One example is extending the work to reinforcement learning (RL) tasks. If ANML helps in that setting it would be important since RL agents naturally encounter sequential learning environments, and are currently held back by all of the challenges that come with them. Another dimension of difficulty that can be tested is when the meta-test distribution is different from the meta-training distribution. If the meta-training distribution is large enough, it would be interesting to test whether ANML can learn to continuously learn in a way that generalizes to entirely new types of tasks that are not in the meta-training set.

Another potentially profitable direction is to hybridize the insights of ANML with other meta-learning techniques. Two promising ones are combining ANML with the style of meta-learning wherein the entire learning algorithm is meta-learned within a recurrent neural network (either in a supervised~\cite{hochreiter2001learning} or RL context, e.g.\ RL$^2$~\cite{duan2016rl} aka Learning to Reinforcement Learn~\cite{wang2016learning}). Another is combining ANML with recent work into meta-learning with differentiable Hebbian learning~\cite{miconi2018differentiable}, including differentiable neuromodulated Hebbian plasticity~\cite{miconi2018backpropamine}. 

Most broadly, the success of OML and ANML underscore the power of meta-learning the solutions to the hardest machine learning problems, such as exploration, safe exploration, generalization, robustness to adversarial examples, and many more. They thus increase our confidence in paradigms like AI-generating algorithms (AI-GAs), which advocate learning as much of the solution as possible~\cite{clune2019ai}. This work focuses on Pillar Two of AI-GAs (meta-learning learning algorithms). The power of the ANML approach should only increase when combined with the other pillars of the AI-GA paradigm, namely architecture search and automatically generating training environments. Architecture search is especially interesting to consider with ANML, because there are so many possible architectures that could be tried and some will likely generate substantial improvements. We could try to investigate that question manually, or we could instead turn to architecture search to discover the answer for us automatically. That will become increasingly attractive as architecture search methods improve.

\section{CONCLUSION}

This work introduces ANML, a method to improve continual learning by directly optimizing for improved continual learning. Here, we have demonstrated the benefits of ANML for reducing catastrophic forgetting. ANML is motivated by the idea that we should meta-learn the solutions to hard problems, instead of manually engineering machine learning solutions for them. Specifically, ANML meta-learns the parameters of a neuromodulatory network that, conditioned on data, gates the activations of a separate prediction network, creating selective activation and, in turn, selective plasticity. The results presented here demonstrate the effectiveness of this approach to reduce the amount of catastrophic forgetting relative to traditional and current state-of-the-art methods at an unprecedented scale, demonstrating continual learning on trajectories of up to 600 sequentially learned classes (over 9,000 SGD updates). 
Although work needs to be done to test how well ANML works on harder challenges, this work provides a promising stepping stone towards improved algorithms for continual learning. It also underscores the value of learning as much of the solution as possible, and thus adds momentum to the growing perspective that we should meta-learn solutions to the grand challenges of AI research. That includes pursuing AI-generating algorithms, which attempt to learn as much as possible in the pursuit of our community's grandest ambition: creating artificial general intelligence.   

\ack 
This work is supported in part by DARPA Lifelong Learning Machines award HR0011-18-2-0018. We thank Hava Siegelmann for her vision in creating that program and for including us in it. We also thank Khurram Javed for providing the OML code and answering clarifying questions about their experiments. Computations were performed on the Vermont Advanced Computing Core (VACC) supported in part by NSF award No. OAC-1827314. We would also like to thank the Vermont Complex Systems Center for assistance, encouragement, and feedback. We are also appreciative of Blake Camp for catching typos in a draft and alerting us to a relevant paper, and to Louis Kirsch for suggesting the addition of a relevant citation. 

\bibliography{ecai.bbl}

\begin{thebibliography}{10}

\bibitem{bear1986modulation}
Mark~F Bear and Wolf Singer, `Modulation of visual cortical plasticity by
  acetylcholine and noradrenaline', {\em Nature}, {\bf 320}(6058),  172,
  (1986).

\bibitem{bengio2015conditional}
Emmanuel Bengio, Pierre-Luc Bacon, Joelle Pineau, and Doina Precup,
  `Conditional computation in neural networks for faster models', {\em arXiv
  preprint arXiv:1511.06297}, (2015).

\bibitem{bengio2013deep}
Yoshua Bengio, `Deep learning of representations: Looking forward', in {\em
  International Conference on Statistical Language and Speech Processing}, pp.
  1--37. Springer, (2013).

\bibitem{ciregan2012multi}
Dan Ciregan, Ueli Meier, and J{\"u}rgen Schmidhuber, `Multi-column deep neural
  networks for image classification', in {\em 2012 IEEE conference on computer
  vision and pattern recognition}, pp. 3642--3649. IEEE, (2012).

\bibitem{clune2019ai}
Jeff Clune, `{AI-GAs}: Ai-generating algorithms, an alternate paradigm for
  producing general artificial intelligence', {\em arXiv preprint
  arXiv:1905.10985}, (2019).

\bibitem{duan2016rl}
Yan Duan, John Schulman, Xi~Chen, Peter~L Bartlett, Ilya Sutskever, and Pieter
  Abbeel, `R{L}$^2$: Fast reinforcement learning via slow reinforcement
  learning', {\em arXiv preprint arXiv:1611.02779}, (2016).

\bibitem{ellefsen2015neural}
Kai~Olav Ellefsen, Jean-Baptiste Mouret, and Jeff Clune, `Neural modularity
  helps organisms evolve to learn new skills without forgetting old skills',
  {\em PLoS computational biology}, {\bf 11}(4),  e1004128, (2015).

\bibitem{fernando2017pathnet}
Chrisantha Fernando, Dylan Banarse, Charles Blundell, Yori Zwols, David Ha,
  Andrei~A Rusu, Alexander Pritzel, and Daan Wierstra, `Pathnet: Evolution
  channels gradient descent in super neural networks', {\em arXiv preprint
  arXiv:1701.08734}, (2017).

\bibitem{finn2017model}
Chelsea Finn, Pieter Abbeel, and Sergey Levine, `Model-agnostic meta-learning
  for fast adaptation of deep networks', in {\em Proceedings of the 34th
  International Conference on Machine Learning-Volume 70}, pp. 1126--1135.
  JMLR. org, (2017).

\bibitem{fix1951discriminatory}
Evelyn Fix and Joseph~L Hodges~Jr, `Discriminatory analysis-nonparametric
  discrimination: consistency properties', Technical report, California Univ
  Berkeley, (1951).

\bibitem{french1991using}
Robert~M French, `Using semi-distributed representations to overcome
  catastrophic forgetting in connectionist networks', in {\em Proceedings of
  the 13th annual cognitive science society conference}, pp. 173--178, (1991).

\bibitem{french1999catastrophic}
Robert~M French, `Catastrophic forgetting in connectionist networks', {\em
  Trends in cognitive sciences}, {\bf 3}(4),  128--135, (1999).

\bibitem{hasselmo1995cholinergic}
Michael~E Hasselmo and Edi Barkai, `Cholinergic modulation of
  activity-dependent synaptic plasticity in the piriform cortex and associative
  memory function in a network biophysical simulation', {\em Journal of
  Neuroscience}, {\bf 15}(10),  6592--6604, (1995).

\bibitem{hochreiter2001learning}
Sepp Hochreiter, A~Steven Younger, and Peter~R Conwell, `Learning to learn
  using gradient descent', in {\em International Conference on Artificial
  Neural Networks}, pp. 87--94. Springer, (2001).

\bibitem{ioffe2015batch}
Sergey Ioffe and Christian Szegedy, `Batch normalization: Accelerating deep
  network training by reducing internal covariate shift', {\em arXiv preprint
  arXiv:1502.03167}, (2015).

\bibitem{ishiguro2003neuromodulated}
Akio Ishiguro, Akinobu Fujii, and Peter~Eggenberger Hotz, `Neuromodulated
  control of bipedal locomotion using a polymorphic cpg circuit', {\em Adaptive
  Behavior}, {\bf 11}(1),  7--17, (2003).

\bibitem{javed2019meta}
Khurram Javed and Martha White, `Meta-learning representations for continual
  learning', {\em Neural Information Processing Systems}, (2019).

\bibitem{kingma2014adam}
Diederik~P Kingma and Jimmy Ba, `Adam: A method for stochastic optimization',
  {\em arXiv preprint arXiv:1412.6980}, (2014).

\bibitem{kirkpatrick2017overcoming}
James Kirkpatrick, Razvan Pascanu, Neil Rabinowitz, Joel Veness, Guillaume
  Desjardins, Andrei~A Rusu, Kieran Milan, John Quan, Tiago Ramalho, Agnieszka
  Grabska-Barwinska, et~al., `Overcoming catastrophic forgetting in neural
  networks', {\em Proceedings of the national academy of sciences}, {\bf
  114}(13),  3521--3526, (2017).

\bibitem{kolouri2019attention}
Soheil Kolouri, Nicholas Ketz, Xinyun Zou, Jeffrey Krichmar, and Praveen Pilly,
  `Attention-based selective plasticity', (2019).

\bibitem{krizhevsky2012imagenet}
Alex Krizhevsky, Ilya Sutskever, and Geoffrey~E Hinton, `Imagenet
  classification with deep convolutional neural networks', in {\em Advances in
  neural information processing systems}, pp. 1097--1105, (2012).

\bibitem{lake2015human}
Brenden~M Lake, Ruslan Salakhutdinov, and Joshua~B Tenenbaum, `Human-level
  concept learning through probabilistic program induction', {\em Science},
  {\bf 350}(6266),  1332--1338, (2015).

\bibitem{lecun1998gradient}
Yann LeCun, L{\'e}on Bottou, Yoshua Bengio, Patrick Haffner, et~al.,
  `Gradient-based learning applied to document recognition', {\em Proceedings
  of the IEEE}, {\bf 86}(11),  2278--2324, (1998).

\bibitem{li2018explicit}
Xuhong Li, Yves Grandvalet, and Franck Davoine, `Explicit inductive bias for
  transfer learning with convolutional networks', {\em arXiv preprint
  arXiv:1802.01483}, (2018).

\bibitem{li2017learning}
Zhizhong Li and Derek Hoiem, `Learning without forgetting', {\em IEEE
  transactions on pattern analysis and machine intelligence}, {\bf 40}(12),
  2935--2947, (2017).

\bibitem{liu2018rotate}
Xialei Liu, Marc Masana, Luis Herranz, Joost Van~de Weijer, Antonio~M Lopez,
  and Andrew~D Bagdanov, `Rotate your networks: Better weight consolidation and
  less catastrophic forgetting', in {\em 2018 24th International Conference on
  Pattern Recognition (ICPR)}, pp. 2262--2268. IEEE, (2018).

\bibitem{lomonaco2017core50}
Vincenzo Lomonaco and Davide Maltoni, `Core50: a new dataset and benchmark for
  continuous object recognition', {\em arXiv preprint arXiv:1705.03550},
  (2017).

\bibitem{maaten2008visualizing}
Laurens van~der Maaten and Geoffrey Hinton, `Visualizing data using t-sne',
  {\em Journal of machine learning research}, {\bf 9}(Nov),  2579--2605,
  (2008).

\bibitem{maclaurin2015gradient}
Dougal Maclaurin, David Duvenaud, and Ryan Adams, `Gradient-based
  hyperparameter optimization through reversible learning', in {\em
  International Conference on Machine Learning}, pp. 2113--2122, (2015).

\bibitem{maltoni2019continuous}
Davide Maltoni and Vincenzo Lomonaco, `Continuous learning in
  single-incremental-task scenarios', {\em Neural Networks}, {\bf 116},
  56--73, (2019).

\bibitem{mann1947test}
Henry~B Mann and Donald~R Whitney, `On a test of whether one of two random
  variables is stochastically larger than the other', {\em The annals of
  mathematical statistics},  50--60, (1947).

\bibitem{MasseE10467}
Nicolas~Y. Masse, Gregory~D. Grant, and David~J. Freedman, `Alleviating
  catastrophic forgetting using context-dependent gating and synaptic
  stabilization', {\em Proceedings of the National Academy of Sciences}, {\bf
  115}(44),  E10467--E10475, (2018).

\bibitem{miconi2018differentiable}
Thomas Miconi, Jeff Clune, and Kenneth~O Stanley, `Differentiable plasticity:
  training plastic neural networks with backpropagation', {\em arXiv preprint
  arXiv:1804.02464}, (2018).

\bibitem{miconi2018backpropamine}
Thomas Miconi, Aditya Rawal, Jeff Clune, and Kenneth~O Stanley, `Backpropamine:
  training self-modifying neural networks with differentiable neuromodulated
  plasticity', (2018).

\bibitem{pham2018efficient}
Hieu Pham, Melody~Y Guan, Barret Zoph, Quoc~V Le, and Jeff Dean, `Efficient
  neural architecture search via parameter sharing', {\em arXiv preprint
  arXiv:1802.03268}, (2018).

\bibitem{rusu2016progressive}
Andrei~A Rusu, Neil~C Rabinowitz, Guillaume Desjardins, Hubert Soyer, James
  Kirkpatrick, Koray Kavukcuoglu, Razvan Pascanu, and Raia Hadsell,
  `Progressive neural networks', {\em arXiv preprint arXiv:1606.04671}, (2016).

\bibitem{schaul2015prioritized}
Tom Schaul, John Quan, Ioannis Antonoglou, and David Silver, `Prioritized
  experience replay', {\em arXiv preprint arXiv:1511.05952}, (2015).

\bibitem{shazeer2017outrageously}
Noam Shazeer, Azalia Mirhoseini, Krzysztof Maziarz, Andy Davis, Quoc Le,
  Geoffrey Hinton, and Jeff Dean, `Outrageously large neural networks: The
  sparsely-gated mixture-of-experts layer', {\em arXiv preprint
  arXiv:1701.06538}, (2017).

\bibitem{soltoggio2008evolutionary}
Andrea Soltoggio, John~A Bullinaria, Claudio Mattiussi, Peter D{\"u}rr, and
  Dario Floreano, `Evolutionary advantages of neuromodulated plasticity in
  dynamic, reward-based scenarios', in {\em Proceedings of the 11th
  international conference on artificial life (Alife XI)}, number CONF, pp.
  569--576. MIT Press, (2008).

\bibitem{soltoggio2018born}
Andrea Soltoggio, Kenneth~O Stanley, and Sebastian Risi, `Born to learn: the
  inspiration, progress, and future of evolved plastic artificial neural
  networks', {\em Neural Networks}, {\bf 108},  48--67, (2018).

\bibitem{stanley2009hypercube}
Kenneth~O Stanley, David~B D'Ambrosio, and Jason Gauci, `A hypercube-based
  encoding for evolving large-scale neural networks', {\em Artificial life},
  {\bf 15}(2),  185--212, (2009).

\bibitem{velez2017diffusion}
Roby Velez and Jeff Clune, `Diffusion-based neuromodulation can eliminate
  catastrophic forgetting in simple neural networks', {\em PloS one}, {\bf
  12}(11),  e0187736, (2017).

\bibitem{wang2016learning}
Jane~X Wang, Zeb Kurth-Nelson, Dhruva Tirumala, Hubert Soyer, Joel~Z Leibo,
  Remi Munos, Charles Blundell, Dharshan Kumaran, and Matt Botvinick, `Learning
  to reinforcement learn', {\em arXiv preprint arXiv:1611.05763}, (2016).

\bibitem{yosinski2014transferable}
Jason Yosinski, Jeff Clune, Yoshua Bengio, and Hod Lipson, `How transferable
  are features in deep neural networks?', in {\em Advances in neural
  information processing systems}, pp. 3320--3328, (2014).

\bibitem{zenke2017continual}
Friedemann Zenke, Ben Poole, and Surya Ganguli, `Continual learning through
  synaptic intelligence', in {\em Proceedings of the 34th International
  Conference on Machine Learning-Volume 70}, pp. 3987--3995. JMLR. org, (2017).

\bibitem{zhang2018top}
Jianming Zhang, Sarah~Adel Bargal, Zhe Lin, Jonathan Brandt, Xiaohui Shen, and
  Stan Sclaroff, `Top-down neural attention by excitation backprop', {\em
  International Journal of Computer Vision}, {\bf 126}(10),  1084--1102,
  (2018).

\bibitem{zoph2016neural}
Barret Zoph and Quoc~V Le, `Neural architecture search with reinforcement
  learning', {\em arXiv preprint arXiv:1611.01578}, (2016).

\end{thebibliography}

\section{SUPPLEMENTARY INFORMATION}
\renewcommand{\thefigure}{S\arabic{figure}}

\begin{figure}[h]
\includegraphics[width=\linewidth]{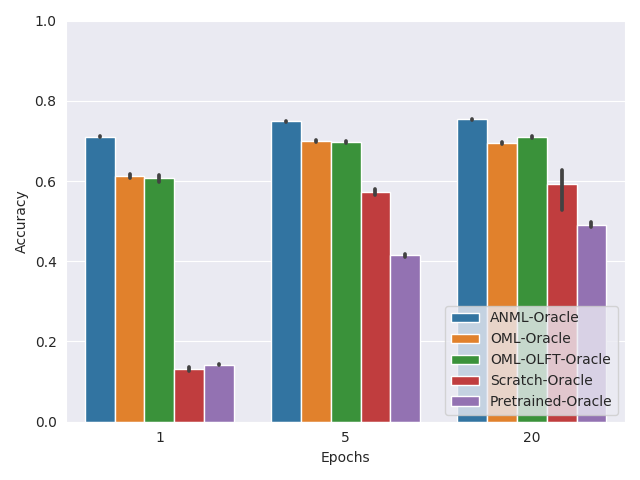}
\caption{
Meta-test testing classification accuracy after 600 classes. Each treatment is trained using the meta-test training data for multiple epochs (x-axis).
}
\label{fig:multiple-epoch-oracles}
\end{figure}

The meta-test training paradigm used here (and borrowed from~\cite{javed2019meta}) fine-tunes models on only a single epoch through the meta-test training data, which contains just 15 images of each class.  While this scenario was specifically chosen to mimic online continual learning, an open question is how the performance advantage of ANML over the other control treatments is affected as additional epochs of training take place (as would be the case in traditional deep-learning settings).  With multiple epochs, we might expect the i.i.d. \textquote{oracle} versions of all the algorithms presented here to increase, as catastrophic forgetting is not an issue in the i.i.d setting, and data/computational limitations are relaxed with multiple epochs.  Indeed, we see that all treatments increase in performance when trained for 20 epochs, compared to training on just a single epoch (all $p<9.67\times10^{-13}$).  However, after training for 20 epochs, the ANML-Oracle still significantly outperforms all other oracles (all $p<3.38\times10^{-12}$), and results in the highest accuracy reported for any treatment throughout this paper, with 75.37\% accuracy when evaluating the network on held-out meta-test testing instances from all 600 classes, after the culmination of sequentially training on the meta-test training instances of those 600 classes.  It is an open, interesting, important research question to figure out exactly why ANML outperforms the controls even in the multi-epoch, i.i.d. setting.

\begin{figure}[h]
\includegraphics[width=\linewidth]{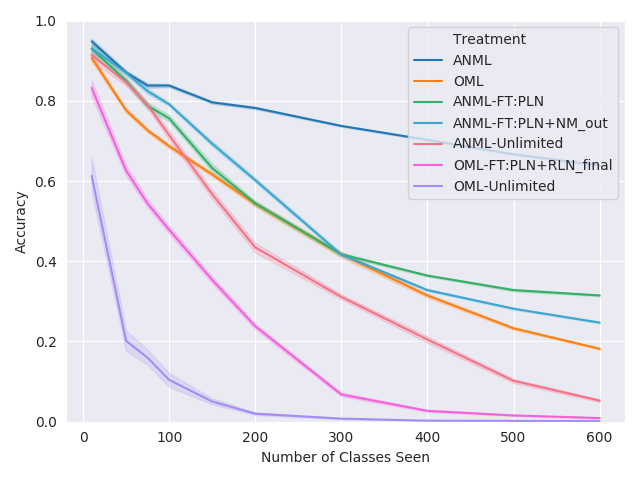}
\caption{
Meta-test test accuracy. Layers that were frozen in the main text are made plastic here. Unfreezing parts of the RLN for the OML treatment resulted in performance ($<1\%$ accuracy) comparable to scratch and pretraining at 600 classes. ANML proved more robust to the degradation of layers that are frozen in the default setup.
}
\label{fig:robustness}
\end{figure}

Complementary to Fig.~\ref{fig:meta-test testing}, which highlights the performance of ANML when a large part of the network is frozen during meta-test training, it is worth reflecting on how performance degrades as parts of the network that were intended to be frozen are made plastic. Fig.~\ref{fig:robustness} compares versions of ANML and OML where different numbers of layers have been allowed to learn during fine-tuning at meta-test training time. For instance, \textit{ANML-Unlimited} means that the prediction learning network (\textit{PLN}) and neuromodulatory network (\textit{NM}; Fig.~\ref{fig:network_architecture}) are fine-tuned at meta-test time (i.e. no layers in either network are frozen).   

Similarly \textit{ANML-FT:PLN} means that the entire prediction network is fine-tuned during meta-test training (while the neuromodulatory network is frozen).  Intermediately, we also examine \textit{ANML-FT:PLN+NM\_out}, which fine tunes the prediction network and the final layer of the NM network (while freezing the convolutional layers of the NM network).  

Naming conventions are only slightly modified for the OML variants in Fig~\ref{fig:robustness}, as OML's \textit{PLN} is defined as its fully connected layers, while its \textit{RLN} is defined as its convolutional layers.  Thus \textit{OML-Unlimited} fine-tunes both the prediction learning network and representation learning network (freezing nothing), which is a treatment designed to investigate how performance of OML drops when all of the meta-learned representation network is subject to fine-tuning.
As a reminder, \textit{OML-FT:PLN+RLN\_final} fine tunes the fully connected layers and also the last convolutional layer.

At 600 classes, both \textit{OML-FT:PLN+RLN\_final} and \textit{OML-Unlimited} fall below 1\% mean classification accuracy on the meta-test test set, representing a 95\% drop in performance relative to the original OML algorithm.

Conversely, \textit{ANML-FT:PLN} (31.5\% accuracy) and \textit{ANML-FT:PLN+NM\_out} (24.7\% accuracy) outperform OML (18.19\% accuracy) at 600 classes ($p = 1.08\times10^{-23}$ and $1.43\times10^{-20}$, respectively).  These two modifications of ANML show a relative drop of 50.6\% and 61.2\%, respectively, from the unmodified ANML performance at 600 classes (63.8\% accuracy).

Even the version of ANML that fine-tunes all parameters in both the \textit{PLN} and \textit{NM} networks at meta-test time, (\textit{ANML-Unlimited}), can protect against forgetting for sequences up to 100 classes before becoming statistically inferior to the unmodified OML algorithm ($p=5.8\times10^{-17}$ at 100 classes). 
On the other hand, both the \textit{OML-Unlimited} and \textit{OML-FT:PLN+RLN\_final}, which allow additional fine-tuning in the OML RLN, perform significantly worse than \textit{ANML-Unlimited} for all lengths of trajectories tested (all $p<4.62\times10^{-14}$). To re-iterate, \textit{ANML-Unlimited} is able to learn up to 100 sequential tasks better than the original OML algorithm, whose RLN is frozen during meta-test training. This result is important and exciting, as it suggests that ANML is able to learn throughout its entire network with much less forgetting than prior methods. The robustness of ANML to perturbations could thus potentially improve transfer to wholly new distributions wherein learning throughout the network may be required, but wherein we still would not want to forget previously learned knowledge catastrophically. 

\begin{figure}[h]
\includegraphics[width=\linewidth]{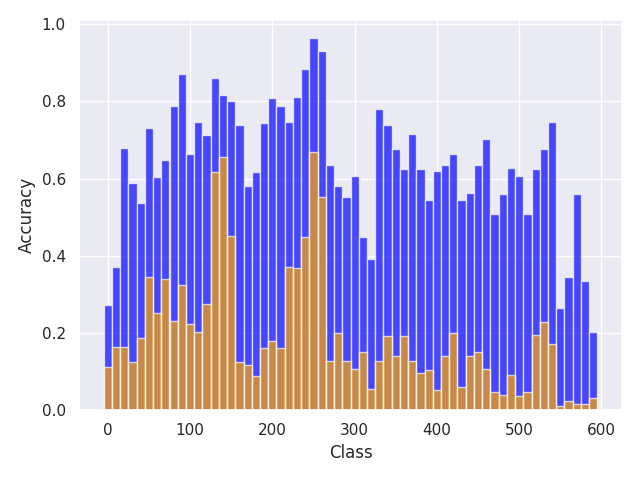}
\caption{
Final accuracy over a 600-task sequence after learning all 600 tasks. After meta-test training on 600 tasks, the model is evaluated on the held-out meta-test test instances.  The accuracy on the test instances is plotted for each class in the order each class was seen during meta-test training. For ease of visualization, sequences of 10 tasks are binned together and their mean across the 10 runs reported in the main text is plotted. 
}
\label{fig:errorDistribution}
\end{figure}

Lastly, to help illuminate the strategy by which each method avoids forgetting, Fig~\ref{fig:errorDistribution} shows the distribution of generalization error for the OML and ANML treatments as a function of the order in which the classes were seen during meta-test training. To aid visualization, each sequence of 10 tasks is binned together. The mean value of each 10-task bin is then computed and plotted as a bar graph.  
Both perform significantly better than chance ($p<7.96\times10^{-6}$) on most classes (99.3\% of classes for ANML and 79.7\% of classes for OML, although performance on most is higher for ANML), including early classes in the 600-task sequence, demonstrating learning without forgetting catastrophically.

\end{document}